\documentclass[letterpaper, 10 pt, conference]{ieeeconf}  

\IEEEoverridecommandlockouts                              

\usepackage{graphicx} 
\usepackage{color}
\usepackage{booktabs}
\usepackage{multirow} 
\usepackage{caption}
\usepackage{lipsum}
\usepackage{tabularx}
\usepackage{float}
\usepackage{stfloats}
\usepackage{makecell}
\usepackage{amsmath,amsfonts}
\usepackage{algorithmic}
\usepackage{url}
\usepackage{cite}
\usepackage[colorlinks=true, linkcolor=blue, citecolor=blue, urlcolor=blue]{hyperref}
\usepackage{adjustbox}
\usepackage[table]{xcolor}

\title{\LARGE \bf
ZING-3D: Zero-shot Incremental 3D Scene Graphs via Vision-Language Models
}

\author{Pranav Saxena$^{1}$, Jimmy Chiun$^{2}$
\thanks{*This work was not supported by any organization}
\thanks{$^{1}$ Pranav Saxena is with Birla Institute of
Technology and Science Pilani, K.K Birla Goa Campus, Goa, India
        {\tt\small f20220257@goa.bits-pilani.ac.in}}%
\thanks{$^{2}$ Jimmy Chiun is with the Department of Mechanical Engineering,
College of Design and Engineering, National University of Singapore.
{
        {\tt\small jimmy.chiun@u.nus.edu}}%
}
}

\begin{document}

\makeatletter
\let\@oldmaketitle\@maketitle
\renewcommand{\@maketitle}{%
    \@oldmaketitle
    \centering
    \vspace{1em}
    \includegraphics[width=\linewidth, height=0.29\textheight,trim={1.1cm 0 0.8cm 0}]{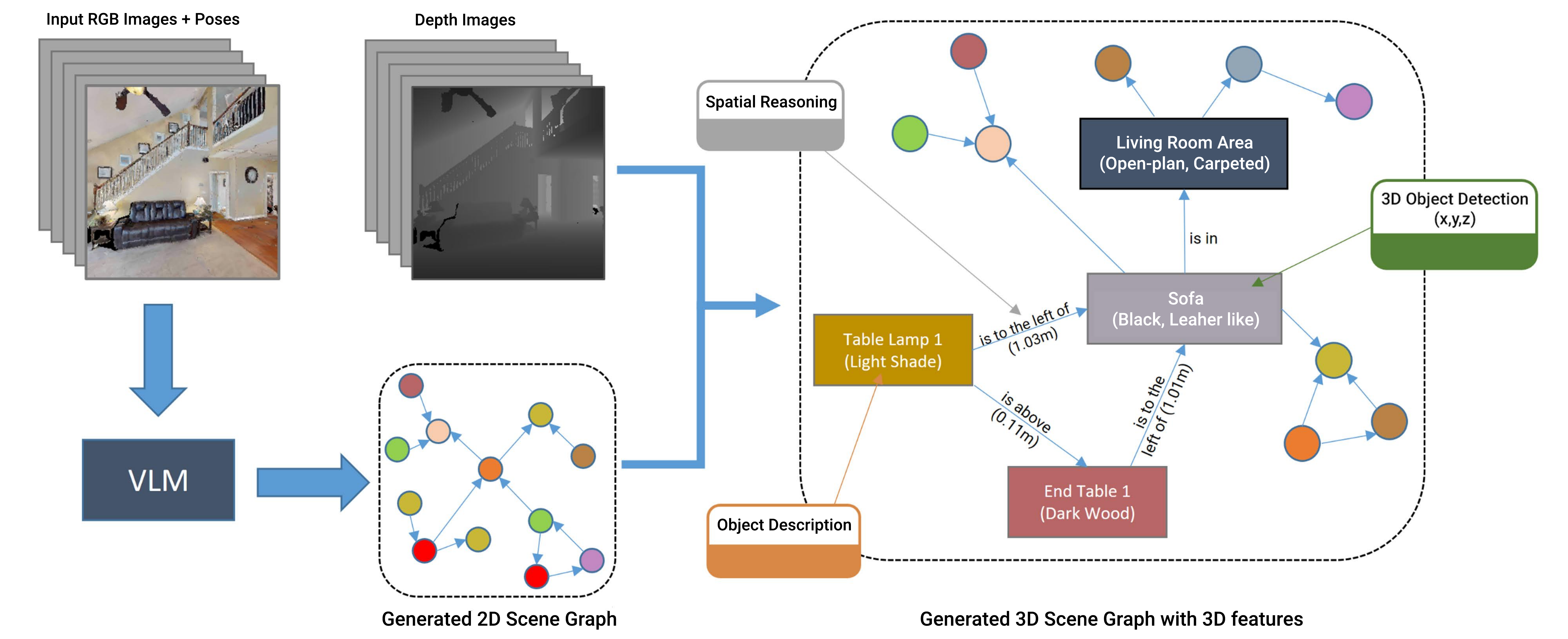}
    \captionof{figure}{
        \textbf{ZING-3D} generates zero-shot open-vocabulary 3D scene graphs and incrementally updates them as a robot navigates an environment. (A) A Large Vision-Language Model outputs a 2D scene graph from a set of posed RGB images. (B) The 2D graph is projected to 3D, incorporating object coordinates, inter-object distances, and spatial relationships. (C) Each node contains object descriptions, associated room type, and object category.}
    
    \label{fig:intro}
    \vspace{-0.15em}
}
\makeatother

\maketitle
\thispagestyle{empty}
\pagestyle{empty}

\begin{abstract}
\setcounter{figure}{1}
Understanding and reasoning about complex 3D environments requires structured scene representations that capture not only objects but also their semantic and spatial relationships. While recent works on 3D scene graph generation have leveraged pretrained VLMs without task-specific fine-tuning, they are largely confined to single-view settings, fail to support incremental updates as new observations arrive and lack explicit geometric grounding in 3D space, all of which are essential for embodied scenarios. In this paper, we propose, \textit{ZING-3D}, a framework that leverages the vast knowledge of pretrained foundation models to enable open-vocabulary recognition and generate a rich semantic representation of the scene in a zero-shot manner while also enabling incremental updates and geometric grounding in 3D space, making it suitable for downstream robotics applications. Our approach leverages VLM reasoning to generate a rich 2D scene graph, which is grounded in 3D using depth information. Nodes represent open-vocabulary objects with features, 3D locations, and semantic context, while edges capture spatial and semantic relations with inter-object distances. Our experiments on scenes from the Replica and HM3D dataset show that ZING-3D is effective at capturing spatial and relational knowledge without the need of task-specific training.

\end{abstract}

\section{INTRODUCTION}

Understanding and reasoning about complex environments requires structured representations that go beyond low-level perception. \textit{Scene Graph Generation (SGG)} has emerged as an effective intermediate representation, mapping raw sensor data or images into a structured graph where objects are nodes and their pairwise relations are edges~\cite{li2017msdn,2,xu2017scene}. Such graphs provide a compact, interpretable, and compositional representation that supports high-level reasoning tasks such as visual question answering, navigation, manipulation, and human-robot interaction. In the context of robotics, the choice of scene representation is critical for downstream planning and decision-making, as robots must construct and maintain such representations online from onboard sensors while operating in dynamic environments. For efficiency and long-term deployment, these representations should be scalable to large environments, open-vocabulary to handle novel objects and relations beyond a fixed training set, and flexible in their level of abstraction to support tasks that range from geometry-rich mobility and manipulation to semantic reasoning and affordance-based task planning. Within this landscape, SGG stands out as a promising approach for bridging raw perception with high-level reasoning, offering interpretability, compositionality, and adaptability for a wide spectrum of real-world applications.~\cite{Werby-RSS-24, gu2024conceptgraphs}.  

Despite their promise, most existing SGG methods are built on supervised training with curated datasets such as Visual Genome or OpenImages~\cite{zellers2018scenegraphs, krishna2016visualgenomeconnectinglanguage, krylov2021openimagesv5text}. While effective in controlled settings, these models face three major limitations. First, they rely on dataset-specific vocabularies, making them brittle when confronted with novel objects or relations in the real world. Second, they primarily operate on single 2D images, which fails to capture spatial consistency or continuity across viewpoints. Third, they are non-incremental, producing one-off graphs without mechanisms to refine or expand the scene representation as an agent explores. These shortcomings make them poorly suited for embodied AI, where agents must incrementally build and refine an understanding of their 3D surroundings in open-world conditions.  

Recent advances in {Vision-Language Models (VLMs) have introduced a paradigm shift, enabling {zero-shot relational reasoning from images. Works such as Open-World SGG~\cite{dutta2025openworldscenegraph} and Pixels-to-Graphs (PGSG)~\cite{li2024pixels} demonstrate that pretrained VLMs, when properly prompted, can generate scene graphs without task-specific fine-tuning. However, these methods remain restricted to single-image 2D settings, limiting their applicability to embodied scenarios. They do not address the challenges of incrementally updating a scene graph as new information becomes available, nor do they integrate geometric grounding in 3D space.  

In this work, we propose ZING-3D, a framework for Incremental 3D Scene Graph Generation using VLMs in a zero-shot manner that addresses the above-mentioned challenges. Our system is designed from the perspective of an embodied agent navigating in a real-world environment. As the agent explores, it captures sequences of egocentric images, which are processed in temporal chunks by a VLM to infer objects and relations in a zero-shot manner. The identified entities are segmented and grounded in the image using an open-vocabulary segmentation model, and subsequently projected into 3D coordinates to produce spatially grounded object instances. As exploration progresses, the scene graph is incrementally updated by merging new observations into a unified global 3D graph that evolves over time.

This pipeline produces a structured, open-vocabulary, and geometrically grounded representation of the environment, that is both semantically rich and spatially aware.  

Our key contributions are:
\begin{enumerate}
    \item \textbf{Zero-Shot Embodied Scene Graph Generation:} Our framework requires no task-specific training, leveraging pretrained VLMs and open-vocabulary object grounding for scalable deployment in unseen environments while being efficient.  
    \item \textbf{Rich Semantic Information:} Our graphs capture detailed semantic and spatial information, with nodes encoding object features, 3D locations, and contextual room type, and edges representing spatial and semantic relationships with precise inter-object distances.
    
    \item \textbf{Incremental Graph Updates:} Our framework incrementally updates the 3D scene graph as the agent explores, producing a temporally consistent, structured representation suitable for downstream robotics tasks.

\end{enumerate}

\section{Related Work}

\subsection{Scene Graph Generation}

Scene Graph Generation (SGG) has emerged as a fundamental task in computer vision, aiming to represent an image as a structured graph of entities and their relationships. Since its introduction by Johnson \cite{Johnson_2015_CVPR}, scene graphs have been widely adopted as intermediate representations for downstream applications, including image retrieval, visual question answering, and image captioning \cite{9031001, schuster2015}. Early approaches primarily focused on predicting pairwise relationships between detected objects, while later works modeled higher-order regularities, such as object–predicate–object motifs, to improve performance \cite{9157682}. Despite progress, SGG remains a challenging problem due to the complexity of natural scenes. A key difficulty is the \textit{long-tailed distribution} of visual relations, where frequent relationships dominate and rare ones are underrepresented. To mitigate this, several methods have explored debiasing strategies, reweighting techniques, and the use of scene context to improve robustness \cite{chen2019,Lu_2021_ICCV}. Another bottleneck is the high annotation cost of scene graph datasets, which has motivated weakly- and semi-supervised approaches \cite{9710723,yang2022}. However, even with these advancements, most approaches remain confined to a \textit{closed-world setting}, where the object and predicate vocabularies are fixed during training.

To move beyond these constraints, recent research has focused on open-vocabulary SGG (OV-SGG). Existing works can be broadly categorized into two directions: (i) \textbf{Open-vocabulary Detection (OvD)}, which predicts relationships among unseen object categories using a fixed predicate vocabulary \cite{gu2022openvocabularyobjectdetectionvision,zang2022open}, and (ii) \textbf{Open-vocabulary Relationships (OvR)}, which recognizes unseen predicates between known objects \cite{he2022openvocabularyscenegraphgeneration,10204761}. More recent efforts attempt to unify both settings (OvD+OvR) \cite{chen2024expandingscenegraphboundaries,salzmann2024scenegraphvitendtoendopenvocabulary}, though they still depend on task-specific supervision and struggle to generalize to unconstrained visual environments. The emergence of large-scale \textit{Vision-Language Models (VLMs)} and advances in zero-shot object detection offer a new paradigm: open-world, training-free SGG. Unlike conventional methods reliant on annotated data, VLMs enable zero-shot generalization to novel objects and relations, making it feasible to construct scene graphs without fine-tuning, where both entities and predicates may be previously unseen.

\subsection{Open-Vocabulary Scene Graph Generation}

\begin{figure*}[t]
    \centering
    \includegraphics[width=1.05\textwidth]{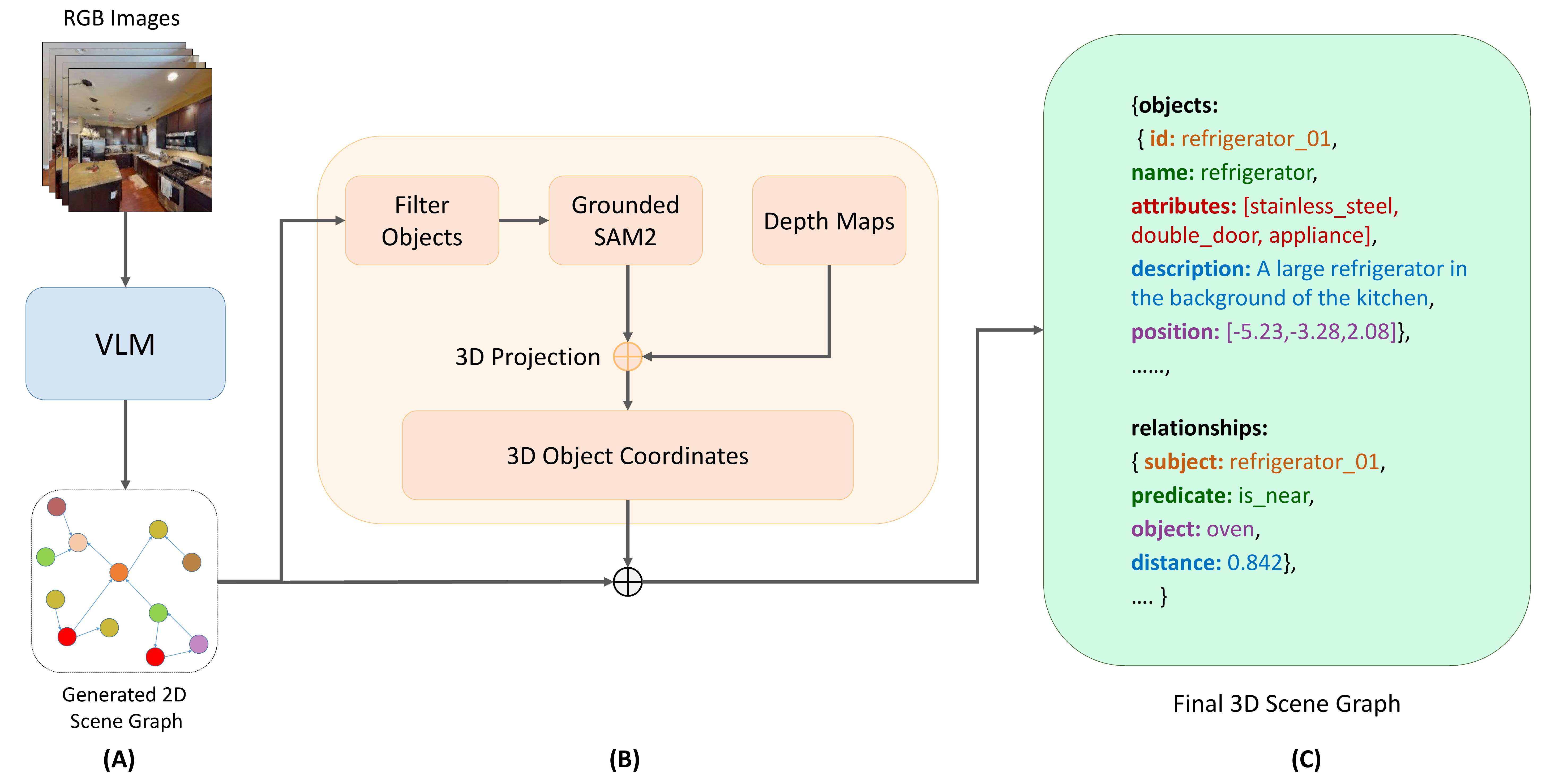}
    \caption{\textbf{Architecture of ZING-3D: }
\textbf{(A)} An open-vocabulary 2D scene graph is built from a sequence of posed RGB images using a large vision-language model (VLM), capturing object descriptors, inter-object relations, room context, and object types. 
\textbf{(B)} The object list is filtered and only relevant relevant furniture and objects are extracted which are then passed to Grounded-SAM2 to generate bounding boxes and corresponding segmentation masks, which are then projected to 3D using depth maps. 
\textbf{(C)} The resulting 3D scene graph provides a structured, comprehensive representation of the scene, suitable for various downstream robotics tasks.
}
    \vspace{-1em}
    \label{fig:arch}
\end{figure*}

OvD methods extend object recognition beyond closed sets by aligning visual region features with pretrained text encoders such as CLIP, enabling prompt-based classification and leveraging weak supervision from captions, web-scale data, or large image–text corpora. While effective at expanding the object label space, these approaches generally retain supervised predicate classifiers, leaving relational generalization underexplored. OvR methods address this by targeting unseen predicates, projecting subject–object pairs and predicate embeddings into shared spaces and employing contrastive objectives or linguistic priors to improve coverage. However, such models remain vulnerable to language bias and spurious co-occurrence, often defaulting to frequent spatial relations.

\subsection{Vision-Language Models (VLMs) for Scene Graph Generation}

Vision-language models (VLMs) pretrained on large image-text corpora have recently reframed scene graph generation (SGG) from a fully supervised, closed-vocabulary problem to a zero-shot, open-world one. Early integrations primarily used VLMs as open-vocabulary object backbones aligning region features to text embeddings to supply nodes to SGG pipelines \cite{lin2022learningobjectlanguagealignmentsopenvocabulary,guo2024regiongptregionunderstandingvision}. Subsequent works extended VLM usage to \emph{predicate} inference via prompt engineering, contrastive scoring of \texttt{(subject, predicate, object)} textual templates, and language-prior regularization. Training-free pipelines combine open-vocab detectors with cross-modal similarity or LLM/VLM reasoning to rank candidate relations without SGG-specific fine-tuning, while structure-aware constraints and calibration mitigate label bias and textual co-occurrence artifacts. Beyond single images, multimodal LLMs/VLMs (e.g., CLIP, BLIP) have been leveraged for compositional generalization, long-tail robustness, and predicate paraphrase handling through prompt ensembles and retrieval-augmented text spaces \cite{Xu_2023_ICCV,wang2024llavasgleveragingscenegraphs}.

However, most prior works are confined to 2D image settings and don't incorporate 3D information. Few approaches integrate geometric grounding or maintain a temporally consistent 3D representation as an agent explores,essential for downstream tasks such as navigation, manipulation, and online reasoning.

\section{Our Approach}

ZING-3D builds a concise, semantically rich representation of a 3D environment. Given a set of posed RGB frames captured incrementally while navigating in an unkown environment, we first utilize a VLM to infer the objects in the scene, associate them across multiple views, capture relationships between the objects and get a 2D scene graph. We then project the scene graph to 3D using depth information to get the complete 3D scene graph, where edges encode distances between objects along with their semantic relationships. This graph is incrementally updated as the robot continues to explore the environment. 

The resulting scene graph, generated in a zero-shot manner, possesses open-vocabulary capabilities, encapsulates object properties and visual features, and supports a wide range of downstream tasks, including object-goal navigation, manipulation, localization, and remapping. The overall approach is illustrated in Fig. \ref{fig:arch}.

\subsection{Open Vocabulary Object Detection}

Traditional models, such as Grounded-SAM2, are constrained by their need for explicit, text-based prompts to identify objects. While these methods demonstrate impressive capabilities in their respective domains, they lack true open-vocabulary detection, as they must be provided with a fixed set of classes to search for, or a specific query for each object of interest. This limitation necessitates manual intervention and is not scalable to the rich, dynamic nature of real-world scenes where a large number of potential objects can exist.

In contrast, our approach leverages recent advancements in large-scale Vision-Language Models (VLMs) to perform open-vocabulary object detection. Rather than being constrained to a closed set of categories, these models allow us to flexibly identify objects described in natural language, even if such objects were not part of any pre-defined ontology. This capability enables true generalization beyond fixed classes, supporting detection of arbitrary objects encountered during scene exploration.

This approach helps in improved reasoning, vision understanding, and grounding capabilities providing robust and scalable detection performance, allowing our scene graph generation framework to operate in a class-agnostic and real open-world manner. As a result, our method not only scales better to diverse environments but also enhances downstream scene graph construction by ensuring that object representations remain comprehensive and adaptive.

\subsection{2D Scene Graph Generation}

\begin{figure}[b]
    \centering
    \vspace{-1.47em}\hspace{-0.8em}\includegraphics[width=0.5\textwidth]{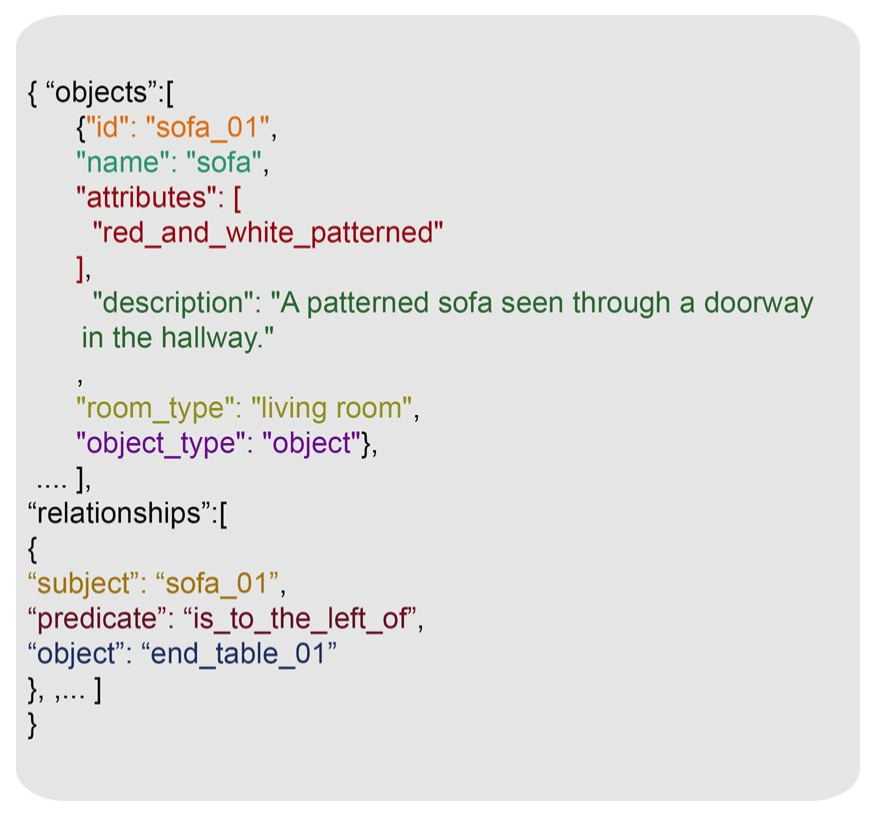}
    \caption{An example of a generated 2D scene graph features.}
    \label{fig:2dsg}
\end{figure}

Once the VLM has identified objects and maintained their associations across multiple frames, the system constructs a 2D scene graph by estimating pairwise relationships between detected entities. Each edge in this graph captures both spatial and semantic interactions, such as “is to the right of,” “is above,” or “is near", that encode the relative positioning of objects within the image plane. Beyond spatial reasoning, the graph also integrates semantic attributes, including the type of environment or room in which an object is located, thereby introducing a hierarchical layer that situates objects within broader contextual groupings. This hierarchical abstraction is particularly important for robotics and embodied AI, as it allows for reasoning not only at the object level (e.g., “chair next to table”) but also at the scene level (e.g., “chair in kitchen”), enabling more structured downstream planning.

In addition to providing a compact and interpretable representation of visual scenes, the 2D scene graph serves as a crucial intermediate step for lifting observations into 3D. It consolidates object features, relational cues, and contextual information into a unified structure that can be projected across frames and viewpoints, making it possible to align multi-view observations into a coherent global representation. An example of a generated 2D scene graph is illustrated in Fig. \ref{fig:2dsg}, showcasing how objects, their relationships, and their scene-level context are jointly represented.

\subsection{3D Projection and Final 3D Scene Graph}

After obtaining a 2D representation of the scene, we extract relevant furniture and objects (excluding trivial classes such as walls or floors) using the VLM. While VLMs are effective for open-vocabulary detection and relation reasoning, they often lack pixel-level precision. To address this, we employ Grounded-SAM2, which provides accurate segmentation masks for each identified object, ensuring precise spatial boundaries and consistent localization across views. These masks are then combined with corresponding depth maps to recover the 3D geometry of each object and align them within the global coordinate frame. This combination of semantic and geometric cues allows the system to accurately associate each object with its real-world position and orientation.

The projection step is carried out using a calibrated point cloud back-projection pipeline. First, we compute the camera intrinsics from the image resolution and field of view. Each valid depth pixel within the segmentation mask is then back-projected into 3D space by mapping image coordinates 
$(u,v)$ with depth $d$ into Cartesian coordinates:

\begin{equation}
X = \frac{(u - c_x)}{f_x} \cdot d, \quad 
Y = d, \quad 
Z = \frac{(v - c_y)}{f_y} \cdot d
\end{equation}

where $(c_x, c_y)$ are the optical centers and $(f_x, f_y)$ are focal lengths. The resulting 3D points are transformed from the camera frame into the global world frame using the robot’s pose (position and orientation), estimated from simulation metadata. For each object, the final 3D centroid is obtained by averaging its valid projected points, yielding a compact and robust location estimate.

This process produces a set of 3D object instances, each associated with a semantic label and global coordinate within the environment. We then fuse this geometric information with the VLM-generated 2D scene graph, enriching it with 3D spatial grounding and metric distances along edges. The result is an open-vocabulary, hierarchically structured 3D scene graph that evolves incrementally as the robot navigates, capturing both semantic relationships and precise geometry in a unified representation.

\subsection{Task-Guided Scene Graph Pruning}
\label{subsec:task_pruning}
To support robots in Vision-Language Navigation (VLN) or object-goal navigation, we perform goal-directed pruning of the 3D scene graph based on the navigation query or target object. The VLM analyzes the scene graph to identify objects most relevant to the task, considering both semantic importance and potential functional interactions with the target. Simultaneously, the robot’s current pose provides spatial context, allowing the system to filter out distant, obstructed, or otherwise inaccessible elements. This results in a focused subset of the graph, removing irrelevant objects and background clutter while preserving key structures and objects critical for navigation.

By concentrating on relevant objects and navigable pathways, this task-specific pruning reduces computational complexity and enables more efficient reasoning during planning. The integration of semantic understanding from the VLM with spatial awareness from the robot ensures that the pruned graph remains informative, actionable, and aligned with the current navigation objective. This compact, contextually aware representation improves situational awareness, supports more accurate decision-making, and enhances the robot’s ability to plan and execute paths effectively toward its goal.

\section{Experiments and Results}

\subsection{Experimental Setup}

\textbf{Environment and System Setup:} We run and evaluate our method on a workstation with one RTX 3090 GPU. 

\textbf{Implementation Details:} The flexible architecture of ZING-3D allows the incorporation of any suitable VLM. Our experiments use Gemini 2.5-Flash as the VLM and Grounded-SAM2 to extract the segmentation masks for 3D projection.

\subsection{Scene Graph Construction}

We evaluate the accuracy of the 3D scene graphs output by the ZING-3D system for scenes in the Replica dataset ~\cite{replica19arxiv} as shown in Table \ref{tab:sg-accuracy}. Since, Replica Dataset has only 1 room per scene, we also evaluate our system on scenes from HM3D dataset. The open-vocabulary nature of our system makes automated evaluation of the quality of nodes and edges in the scene graph challenging. We instead evaluate the scene graph by using human evalutation. For each node, we compute precision as the fraction of nodes for which human evaluation considers the node caption correct. 

We also report the number of
room precision retrieved by each variant by asking evaluators whether they deem each room type a valid one for the corresponding object. ZING-3D identifies a number of valid objects in each scene, and
incurs only a small number (0-1) of duplicate detections. The
node labels are accurate about 97\%\ of the time when tested on the Replica dataset and about 96\%\ on the HM3D dataset scenes. The edges (spatial relationships) are labeled with a high degree of accuracy (96-98\%\ on average).

\begin{table}[!t]
    \centering
    \adjustbox{max width=\linewidth}{%
    
    \begin{tabular}{lccccc}
        \toprule
        & scene & node prec. & room prec. & duplicates & edge prec. \\
        \midrule
        \multirow{8}{*}{Replica}
        & room0 & 0.98 & - & 0 & 0.97 \\
        & room1 & 0.97 & - & 0 & 0.96 \\
        & room2 & 0.97 & - & 0 & 0.98 \\
        & office0 & 0.98 & - & 1 & 1.0  \\
        & office1 & 0.96 & - & 0 & 0.98 \\
        & office2 & 0.97 & - & 0 & 0.95 \\
        & office3 & 0.95 & - & 1 & 0.92 \\
        & \cellcolor{gray!25} \textbf{Average} & \cellcolor{gray!25} \textbf{0.97} & \cellcolor{gray!25} - & \cellcolor{gray!25} - & \cellcolor{gray!25} \textbf{0.96} \\
        \midrule
        \multirow{5}{*}{HM3D}
        & 00801 & 0.95 & 0.93 & 1 & 0.97 \\
        & 00820 & 0.96 & 1.0 & 0 & 0.96 \\
        & 00877 & 0.97 & 1.0 & 0 & 1.0 \\
        & 00894 & 0.96 & 0.97 & 0 & 0.98  \\
        & \cellcolor{gray!25} \textbf{Average} & \cellcolor{gray!25} \textbf{0.96} & \cellcolor{gray!25} - & \cellcolor{gray!25} - & \cellcolor{gray!25} \textbf{0.98} \\
        \bottomrule
    \end{tabular}
    } %
    \caption{\textbf{Accuracy of constructed scene graphs}: node precision indicates the accuracy of the label for each node (as measured by human evaluation); room precision indicates the accuracy of each estimated corresponding room type for an object (human evaluation); duplicates are the number of redundant detections; edge precision indicates the accuracy of each estimated spatial relationship (again, determined by human evaluation)}
    \label{tab:sg-accuracy}
    \vspace{-2em}
\end{table}

\subsection{Vision Language Navigation Queries}

As discussed in Subsection~\ref{subsec:task_pruning}, to evaluate ZING-3D in an embodied AI setup performing Vision-Language Navigation, we apply goal-directed pruning of the 3D scene graph based on the language query. For each navigation task, the VLM reasons over the generated scene graph to identify the most relevant goal object as well as its neighboring objects, with a maximum of eight selected per query. This pruning produces a compact, task-specific scene graph that retains only the objects most critical for navigation. A representative output of this pruning process is shown in Fig.~\ref{fig:pruning}.

\begin{figure}[H]
    \centering
    \includegraphics[width=0.45\textwidth]{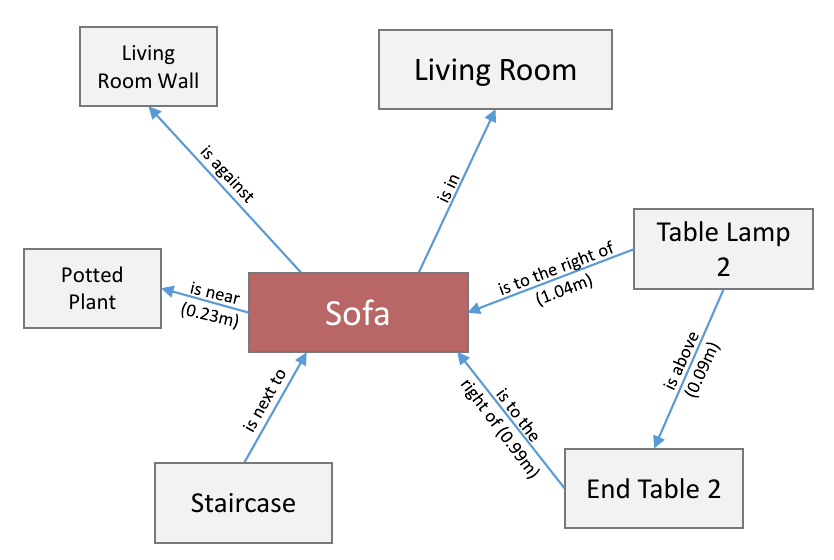}
    \caption{Example pruned scene graph for VLN Query - "Go near the sofa".\vspace{-1.47em}}
    \label{fig:pruning}
\end{figure}

\subsection{Ablation Study}

In this section, we conduct an ablation study to compare the performance of our framework under different design choices. Specifically, we evaluate how variations in the Vision-Language Models (VLMs) used in the system affect the quality of object detection, relationship inference, and 3D scene graph generation.

\begin{table}[H]
    \centering
    \renewcommand{\arraystretch}{0.95}
    \begin{adjustbox}{max width=\linewidth}
    \begin{tabular}{l *{4}{c}}
        \toprule
        Design Choices &  Time Taken(s) & node prec & valid objects\% &edge prec. \\
        \midrule
        Gemini 2.5-flash-lite     & 4.21 & 0.93 & 91 & 0.93 \\
        Gemini 2.5-flash              & 43.13 & 0.96 & 93  & 0.94 \\
        Qwen2.5-VL-7B           & 280 & 0.83 & 85 & 0.88 \\
        \bottomrule
    \end{tabular}
    \end{adjustbox}
    \caption{Comparison of performance across different VLMs. Metrics reported include the time taken to generate the intermediate 2D scene graph from a set of 10 posed observations, node precision, number of valid objects, and edge precision, averaged over 4 evaluated scenes in Habitat-Sim, similar to the setup shown in Table~\ref{tab:sg-accuracy}.}
    \label{tab:sr}
\end{table}

\section{CONCLUSION}

In this work, we presented ZING-3D, a framework for incremental 3D scene graph generation using vision-language models in embodied AI environments in a zero-shot manner. By combining VLM-based object and relation reasoning with accurate 3D projection and incremental fusion, our approach constructs open-vocabulary, spatially grounded 3D scene graphs that evolve as the agent navigates. By associating objects with their room types, ZING-3D further forms a hierarchical 3D scene graph capturing inter and intra-room relations. Additionally, Task-guided pruning produces compact, goal-specific scene graphs, enabling efficient reasoning for Vision-Language Navigation and object-goal tasks. Our experiments show that ZING-3D effectively integrates semantic understanding, geometric grounding, and hierarchical structuring, resulting in a robust and interpretable representation of complex environments

This work highlights the potential of leveraging large VLMs for structured, real-world scene understanding and lays the foundation for future research on fully autonomous agents capable of open-world reasoning, and dynamic environment interaction.

\addtolength{\textheight}{-12cm}   

\bibliographystyle{unsrt}
\bibliography{references}

\end{document}